\documentclass{article}
\usepackage{ijcai20}

\usepackage{times}
\usepackage{soul}
\usepackage{url}
\usepackage[hidelinks]{hyperref}
\usepackage[utf8]{inputenc}
\usepackage[small]{caption}
\usepackage{graphicx}
\usepackage{amsmath}
\usepackage{amsthm}
\usepackage{booktabs}
\usepackage{algorithm}
\usepackage{algorithmic}
\title{Hinting Semantic Parsing with Statistical Word Sense Disambiguation}

\author{
Ritwik Bose
\and
Siddharth Vashishtha \and
James Allen
\affiliations
University of Rochester\\
\emails 
\{rbose, svashis3, james\}@cs.rochester.edu,
}

\begin{document}

\maketitle

\begin{abstract}
The task of Semantic Parsing can be approximated as a transformation of an utterance into a logical form graph where edges represent semantic roles and nodes represent word senses. The resulting representation should be capture the meaning of the utterance and be suitable for reasoning. Word senses and semantic roles are interdependent, meaning errors in assigning word senses can cause errors in assigning semantic roles and vice versa. While statistical approaches to word sense disambiguation outperform logical, rule-based semantic parsers for raw word sense assignment, these statistical word sense disambiguation systems do not produce the rich role structure or detailed semantic representation of the input. In this work, we provide hints from a statistical WSD system to guide a logical semantic parser to produce better semantic type assignments while maintaining the soundness of the resulting logical forms.  We observe an improvement of up to 10.5\% in F-score, however we find that this improvement comes at a cost to the structural integrity of the parse
\end{abstract}

\section{Introduction}

In this work, we improve the sense disambiguation of a logical semantic parser, The TRIPS Parser \cite{allen2008deep}, by integrating advice from a statistical Word Sense Disambiguation (WSD) system, SupWSD \cite{papandrea2017supwsd}. We provide sense decisions from SupWSD to the TRIPS parser to augment existing heuristics and try to keep the correct sense in the search space.

Given word sense advice from SupWSD \cite{papandrea2017supwsd}, a statistical word sense disambiguation system, we improve the quality of logical forms produced by the TRIPS semantic parser.  We provide two types of hints to the parser: \textit{Prehinting} primes the parser to consider certain senses at the beginning of the parsing process while \textit{progressive} hinting continually directs the parser towards certain senses throughout the process.  

\section{Background and Motivation}
Semantic parsing generates logical forms to represent the meaning of natural language utterances.  The resulting logical form is a graph relating nodes, representing word senses, with edges, representing semantic roles.  This graph can be used to performing reasoning for a number of downstream tasks. 
Errors in semantic parsing often manifest as a combination of incorrect senses and roles due to incorrect semantic restrictions being enforced.  By the same token, correcting individual word sense errors can rectify incorrect senses and roles by enforcing the correct semantic restrictions.

%
\begin{figure}
\includegraphics[width=\columnwidth]{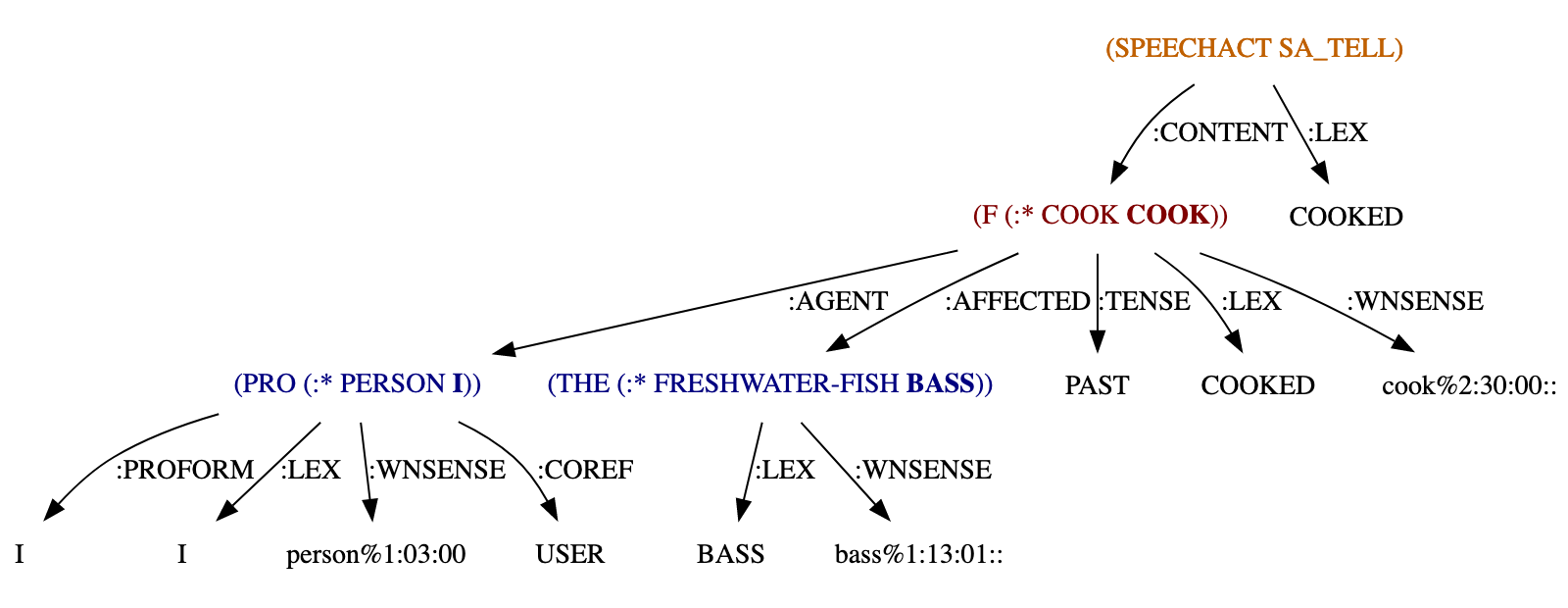}
\caption{The TRIPS parser output for ``I cooked the bass."}
\label{fig:grilledbass}
\end{figure}

\begin{figure}
\includegraphics[width=\columnwidth]{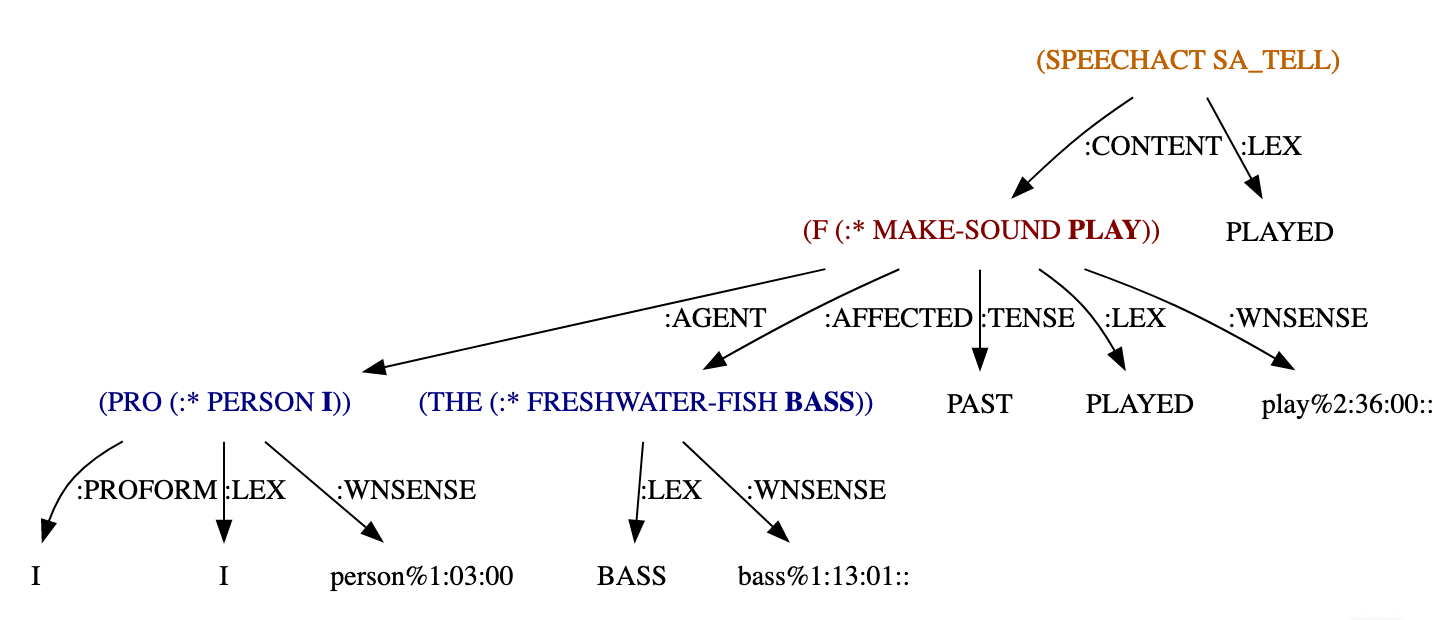}
\caption{The TRIPS parser output for ``I played the bass."}
\label{fig:basssound}
\end{figure}

The TRIPS parser uses a handcrafted grammar with features paired with a (handcrafted) ontology and lexicon.  Broad-coverage is achieved by mapping to larger scale resources, such as WordNet \cite{fellbaum1998semantic} and application of heuristics to prune the search space.  A large number of word sense errors in the TRIPS parser occur when the likelihood of the correct sense falls below a threshold and is pruned from the search space. We expect this to improve the quality of the semantic parses by the TRIPS-parser.

A highly simplistic view of semantic parsing is to consider it as an assignment of semantic types and semantic roles.  Predicates such as \textit{ont::cook} and \textit{ont::make-sound} have argument templates with selectional restrictions on roles.

Figures \ref{fig:grilledbass} and \ref{fig:basssound} show potential parses for the sentences 

\begin{enumerate}
\item ``I cooked the \textbf{bass}" 
\item ``I played the \textbf{bass}"  
\end{enumerate}

We see the parse for the former is correct while the latter mislabels \textbf{bass} as \texttt{ont::freshwater-fish} instead of the correct type, \texttt{ont::musical-instrument}.  The restriction for the \texttt{:affected} role allows for children of \texttt{ont::phys-obj} of which both fish and guitars are instances.  Strict restrictions of \texttt{:affected} to \texttt{musical-instrument} would prevent correct parses of sentences like ``I clanked the chain". The combinatorial restriction produced by \textbf{play} and \textbf{bass}, however, produces the necessary sense. However, it is insufficient to construct a semantic graph and simply replace the sense labels with those determined from a WSD system, as this may lead to semantic role violations in the resulting structure.  Instead, we seek to integrate word sense preferences into the restriction based parsing process in an attempt to settle on a parse that maintains semantic consistency as its primary concern.

In previous work, Skeleton-score \cite{bose2018skeletonscore} guided the TRIPS parser with simple semantic predicates, consisting of fragments of gold annotated parses.  In similar fashion to this work, constituents were boosted for agreement with fragments and penalized for disagreement.  Due to sparse data, it was determined to be beneficial to limit negative evidence and only act on positive evidence.  Strongly positive evidence indicates a match whereas strongly negative evidence may indicate a gap in the data. The Skeleton-Score system achieved a 3.1\% improvement on F-score, scored using SMatch \cite{cai2013smatch}.


\section{Background}

\subsection{Related Work}

Several models of semantic representation have resulted in a variety of different types of semantic parsers.  AMR parsers, such as \cite{peng2015synchronous}, \cite{zhang-etal-2019-amr}, or \cite{flanigan-etal-2016-generation} use a substantial amount of manually annotated structures.  Senses in AMR are drawn from PropBank \cite{propbank}.  \cite{cwang-brandeis} notes that close to two thirds of tokens in AMR annotations are not predicates.  However, as \cite{SONG-BLEU-2019} observes, evaluation for AMR (performed by SMATCH \cite{cai2013smatch}), strongly favors structural matches, further downplaying the importance of word sense disambiguation.  \cite{transductive2019} presents a transductive model for creating Universal Decompositional Semantics representations.  These representations contain both structural and semantic information, including word sense information derived from decomposed WordNet sense annotations \cite{White2016}.  Parsers like FRED \cite{fred} and KNeWS \cite{knews}, both based on Boxer \cite{boxer} provide WSD output along with the generated RDF structure.  However, the WSD output does not influence the argument structure of the final output.

\subsection{Word Sense Disambiguation} 
WSD systems can be mainly divided into three parts: (i) supervised systems, (ii) knowledge-based unsupervised systems, and (iii) fully unsupervised systems. 
Supervised systems rely on annotated corpora such as SEMCOR ~\cite{miller1994using} and DSO corpus ~\cite{ng1996integrating} which have correct word-senses identified in them, mostly hand labelled. These corpuses can then be used to predict the correct word-sense using some machine learning approaches such as Support Vector Machines ~\cite{lee2004supervised}, kNN classifiers~\cite{ng1996integrating,stevenson2001interaction} etc. The supervised approach has proven to be quite successful in many SemEval workshops ~\cite{palmer2001english,snyder2004english,pradhan2007semeval}. In recent years, people have also explored incorporating word-embeddings such as Word2Vec ~\cite{mikolov2013distributed} which can be learned from a large source of unlabeled text and then can be used as features in a WSD algorithm. A recent paper by \cite{kumar-etal-2019-zero} uses continuous embedding space to tackle unseen words in a corpus and achieves state-of-the-art on WSD performance.

In this work, we use a supervised-system SupWSD ~\cite{papandrea2017supwsd} which provides a probability distribution over WordNet senses for a given sentence. The system uses an SVM classifier to predict the word senses and its features include various linguistics properties such as POS tags, syntactic relations, local collocation, word-embeddings, and information about surrounding words. We show that combining the knowledge of a statistical WSD with a semantic parser results into an improved performance of the semantic parser.  While it is noted that SupWSD does not provide state of the art WSD, the results are competitive and the goal of this work is to examine the impact of including WSD information.

\subsection{The TRIPS Parser}

The TRIPS parser is a best-first bottom-up chart-parser with a hand-built, lexicalized context-free grammar.  A core semantic lexicon produces lexical entries which unify syntactic features and templates with semantic types drawn from an ontology.  Constituents are constructed bottom-up and scored by several heuristics for likelihood.  Broad coverage for lexical entries is achieved by hand-mapping the multiple inheritance WordNet 3.0 hierarchy \cite{fellbaum1998semantic} to the single inheritance TRIPS ontology.

\subsubsection{The TRIPS Ontology} 

\begin{figure}
\includegraphics[width=\columnwidth]{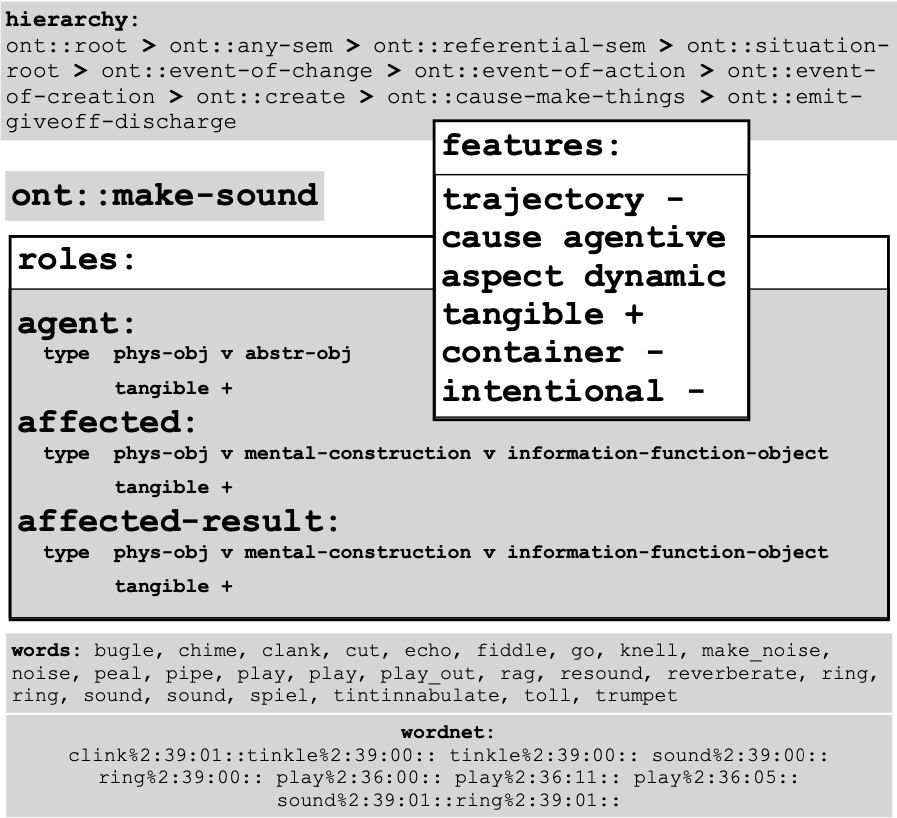}
\caption{Ontology entry for \texttt{ont::make-sound}.}
\label{fig:ontgraph}
\end{figure}

The ontology is a hand-crafted single-inheritance hierarchy where nodes specialize or override hierarchical features and argument templates from their parent.  Each sense is related to its parent and children in both syntax and semantics, so word instances which share a sense also share grammatical constructions.  Each sense specifies a set of roles with type and feature restrictions. Hence, the graphical structure of logical forms generated by the parser is as much determined by the word senses provided by TRIPS types as semantic roles.

In order to leverage the broad lexical coverage provided by WordNet, the TRIPS ontology also contains manual mappings from WordNet synsets to ontology types.  Using a simple subsumption algorithm, sub-trees of the WordNet hierarchy are assigned to TRIPS senses.

A word, $w$ belonging to a synset $s_k$, can be assigned a TRIPS sense, $T_n$ if there is a sequence of synsets, $\{s_k ... s_{k+i}\}$ where $s_j$ is a hyponym of $s_{j+1}$, $s_{k+i}$ maps to $T_n$ and none of $\{s_k ... s_{k+i-1}\}$ map to another TRIPS sense.

\subsubsection{The TRIPS Lexicon} 
The lexicon contains definitions for over 7000 words generated from a number of sources, with a core of 645 handcrafted definitions.  A definition consists of a syntactic template and a semantic type from the ontology, each with corresponding features.  The combination of syntactic and semantic information produce restrictions and preferences for semantic roles to link the graph. 

\subsubsection{Building Lexical Entries}
The parser starts by extracting lexical and grammatical features from the sentence to constrain and identify plausible word definitions and syntactic templates. Since the WordNet mappings to the ontology are not included in the lexicon, it is necessary to generate lexical entries for them as well.  Lexical information, such as subcategorization and morphology, is used to constrain possible syntactic templates.  The parser retrieves candidate senses from the ontology by way of WordNet in addition to the standard lexicon lookup.  To attach syntactic templates to the candidate senses, the parser examines existing lexical definitions using the sense which share lexical information.  Candidate entries are scored using a variety of heuristics and poorly scored entries are pruned off.  A lexical entry in the TRIPS parser can be represented as a tuple $(F, S, T)$, where $F$ represents lexical features, $S$ represents a syntactic template, and $T$ represents an ontology type.

\subsubsection{The Chart}
In order to keep the search space manageable, the parser continuously applies heuristics to constituents on the chart, considering only those constituents with high scores.  Constituents are also constantly pruned from the chart.  The parser terminates when it has found a sufficient spanning parse or determines the likelihood of finding one to be below a threshold.  We integrate progressive hinting here by modifying the score of a constituent as it is considered by the chart.

\section{Methodology}

In this work, we use a supervised-system SupWSD ~\cite{papandrea2017supwsd} which provides a probability distribution over WordNet senses for a given sentence. The system uses an SVM classifier to predict the word senses and its features include various linguistics properties such as POS tags, syntactic relations, local collocation, word-embeddings, and information about surrounding words. We show that combining the knowledge of a statistical WSD with a semantic parser results into an improved performance of the semantic parser.  While it is noted that SupWSD does not provide SOTA WSD, the results are competitive and the goal of this work is to examine the impact of including WSD information.

Word sense advice consists of a word, sense, and score over a token span.  The on receiving a new sentence, the parser receives a message indicating SupWSD's sense preferences for each span.  This contains all the information required to perform either type of hinting.  The advice module converts WordNet senses output by SupWSD into a TRIPS ontology type. This is to ensure minimal interaction between the word sense advice and other internal heuristics in the parser.

\section{Implementation Details}   
\subsection{Tokenization and unification} We find that the TRIPS parser, SupWSD, and Semeval2015 data each left to their own devices produce distinct tokenizations of words.  Attempting to enforce foreign tokenization in the TRIPS parser results in a drastic reduction of parsing quality.  Instead, we allow the systems to tokenize as they please and then apply a sequence of heuristics to unify tokens when transmitting information.  In general, we claim a match between two elements if their spans intersect and base form of the word is an exact match.  For any remaining unmatched elements, we claim matches if the words are identical.

\subsection{Advice transformation} 
Advice received from SupWSD comes in the form of a map of senses and scores.  WordNet senses are mapped to TRIPS types using subsumption and the resulting trips type receives the probability mass of its parent as a score.  When the parser requests advice, the word sense module selects the highest scoring item or items and sends them to the parser.    

\subsubsection{Prehinting} 
Prehinting ensures the parser adds a lexical entry with the given sense to its chart. Prehinted senses start on the chart with a score of 1, which generally ensures they are considered highly early in the parsing process.

\subsubsection{Progressive Hinting}
Progressive hinting is performed as constituents are expanded on the chart.  At each stage, constituents are pruned from the chart performed to ensure the search space remains feasible. The goal of progressive hinting is to subtly augment the score of constituents containing senses advised by SupWSD without overwhelming the existing grammar and heuristics.  Augmenting the constituent's score by a small factor every round decreases but does not eliminate the possibility of being pruned.

Progressive hints are specified as a tuple $(w, i, t, s)$, where $w$ is the base-form of a word, $i$ is an interval representing the span of the word, $t$ is a trips type and $s$ is a score representing the confidence associated with the hint.  The parser stores all hints for a sentence in the advice-map at the start of parsing.  As constituents are selected, the advice-map is queried for word and span. For a word/span query, $(w_c, i_c, t_c)$ from the constituent, if $w = w_c$, $i \cap i_c \neq \emptyset$ and $t \in {ancestors}(t_c)$, we augment the constituent score by $s$.  Specifically, 
$${augment}(s_c, s, \alpha) = \alpha * (s_c * s) + (1 - \alpha), 0 \leq \alpha \leq 1$$  The final score is computed recursively as the average of the augment of the root constituent and the score of its children.

We distinguish factors by semantic boundaries in the underlying resource by partitioning the TRIPS ontology at nodes with a semantic structure different to their parent. The resulting partition has the property that all members of a partition are closely related by semantic content.  It follows that for two nodes to fall in different partitions, they have a specific, quantifiable difference in meaning which can directly impact parse structure.  We apply this concept to determine the severity of sense deviances.  If a predicted sense falls in the same semantic factor as their gold tags, we have evidence which suggests that the predicted sense will not cause structural damage to the rest of the parse.

\section{Evaluation}

In order to evaluate the system, we run the variants of TRIPS parser on SemEval2015 \cite{moro-navigli-2015-semeval}\footnote{We utilized the XML-formatted SemEval2015 data from \cite{raganato2017word}}.  We compare the sense tagging produced by the parser to that produced by SupWSD.  The purpose of these experiments is to determine whether or not the information from the WSD system can be transferred to the semantic parsing domain, not to try to elicit an improvement on WSD.  

\subsection{Fragmentation}

In the case that TRIPS cannot produce a feasible spanning parse without extensive role violations, the TRIPS parser returns best-effort fragments.  Many of the sentences in the test set are quite grammatically complicated and stylized resulting in extensive fragmentation.  We analyze the impact of the provided hints on sentence fragmentation. Semeval2015 is annotated with WordNet senses which draw different semantic and syntactic distinctions to TRIPS senses.  Providing a WordNet derived sense as the only option for a particular word may cause a sentence to fragment due to the lack of adequate semantic templates.  In addition, in an attempt to save parse cohesion, the TRIPS parser might choose to tag words with the fallback type \texttt{ont::referential-sem}.

\subsection{Semantic Factor Agreement}

We produce a factorized version of the ontology by partitioning the TRIPS ontology at nodes which have a semantic structure different to their parent. The resulting partition has the property that all members of a partition have the same role structure and restrictions and could therefore be substituted with each other for a structurally valid parse.  It follows that for two nodes to fall in different partitions, they have a specific, quantifiable difference in meaning which can directly impact parse structure.  We apply this concept to determine the severity of sense deviances.  For a word instance, if the predicted sense falls in the same semantic factor as the gold tag, the predicted sense will not cause a significant structural change to the rest of the parse than if the gold tag were applied.

\subsection{Experiments}

We produce variants of the parser with no hinting (\texttt{plain}), pre- and progressive hinting, and a hybrid system combining both.  Additionally, we produce a \texttt{fixed} variant of the parser which guarantees tagged words will receive the best sense determined by SupWSD.

\begin{table}[htb]
\centering
\begin{tabular}{l|c|c|c|c|c}
\toprule
\multicolumn{1}{c}{\textbf{System}} & 
\multicolumn{1}{c}{\textbf{F-Score}} & 
\multicolumn{1}{c}{\textbf{WuPalmer}} & 
\multicolumn{1}{c}{\textbf{SemFac}} &
\multicolumn{1}{c}{\textbf{Frag}} \\
\midrule
Plain & 48.2 & 76.7  & 54.8 & 75 \\
\midrule
Pre & 47.9  & 75.3 & 54.4 & 67 \\ 
\midrule
Prog & 52.0 & 78.9 & 60.4 & 72 \\
\midrule
Comb & 53.2 & 79.1 & 61.8 & 65 \\
\midrule
Fixed & 58.7 & 81.8 & 68.5 & 97 \\
\midrule
SupWSD & 67.3 & 81.8 & 70.8 & - \\ 
\bottomrule
\end{tabular}
\caption{\label{tab:wup_table}Results from evaluation on SemEval-2015 \protect\cite{raganato2017word}.  We compare accuracy, Wu-Palmer and semantic factor agreement.}
\end{table}
We report three metrics to evaluate the performance of the semantic parser -- (i) accuracy (exact sense agreement) (ii) Mean Wu-Palmer similarity \cite{wu1994verbs} and (iii) Mean SemFac (computed as Wu-Palmer over the factorized ontology described above).  We find that sense decisions provided by SupWSD are able to substantially improve the decisions made by the TRIPS parser.  Correct decisions from SupWSD improve the general accuracy of the parser while incorrect decisions from SupWSD do not cause cascading errors.

Table \ref{tab:wup_table} shows that the \textit{plain} parser significantly under-performs compared to SupWSD. However, we see improvements across all metrics in both the \textit{pre} and \textit{prog} strategies of hinting the parser. When the strategies are combined we see a significant improvement across all metrics, with 10.5\% improvement to F-score and 14.3\% improvement in semantic factor accuracy.  However, the largest improvement comes with a large increase in fragmentation.  The more modest improvement produced by the pre-hinting and combined strategies reduce fragmentation because the improvements elicited from these systems are the result of better and more options in the parser.

\section{Discussion and future work}

As is evidenced by the relative success of strategies including prehinting, a major cause of word sense level errors in the TRIPS parser is due to the correct sense not being available on to the chart at the time of parsing.  Due to the relative orthogonality between type assignments and semantic structure, the problem of such cascading errors is not as pressing in approaches such as AMR.

In order to offset the errors induced by incompatible senses, it may be useful to incorporate sub-sense level semantic information in determining possible word senses.  UCCA sense annotations with decomposed WordNet senses \cite{White2016} may align well with handcrafted semantic features in the TRIPS ontology providing more fine-grained senses.  Alternatively, ULF \cite{gene-2019} provides an underspecified semantic annotation with compatible argument structures to the TRIPS annotation scheme.  Structural hints inferred from ULF may be used to filter candidates senses or prefer constituents on the TRIPS chart which exhibit similar connections to those in the ULF parse in a similar manner to the progressive hinter.

\section{Conclusion}

In this work, we use SupWSD, a statistical WSD system, to advise the TRIPS parser, a logical semantic parser.  Our initial attempts at providing advice both to initialize and guide the parser elicit significant improvements in the word sense decisions output by the parser.  We define and use \textit{SemFac} similarity over the TRIPS ontology to examine impact of modified senses in the parser output and determine that the resulting senses do not introduce significant new conflicts in semantic restrictions.  We find that introducing word sense advice can elicit greater improvements in the quality of semantic parsing, but relying too heavily on WSD output results in an increase parse in fragmentation.  This is because as senses from SupWSD are incorporated into the parse, due to differences in the way that WordNet and TRIPS organise senses, semantic restrictions become harder to fulfill.

\bibliographystyle{ijcai20}
\bibliography{bibfile1}

\end{document}